\documentclass[sigconf, nonacm]{acmart}
\AtBeginDocument{%
  }

\setcopyright{acmlicensed}
\copyrightyear{2025}
\acmYear{2025}
\acmDOI{XXXXXXX.XXXXXXX}
\acmConference[Conference acronym 'XX]{Make sure to enter the correct
  conference title from your rights confirmation email}{June 03--05,
  2025}{Woodstock, NY}
\acmISBN{978-1-4503-XXXX-X/2025/06}

\usepackage{multirow}
\usepackage[most]{tcolorbox}

\begin{document}

\title{GeoFlow: Agentic Workflow Automation for Geospatial Tasks}

\author{Amulya Bhattaram}
\email{abhattaram@utexas.edu}
\authornote{Equal contribution; alphabetical order.\\Preprint. Paper accepted to ACM SIGSPATIAL 2025.}
\affiliation{%
  \institution{The University of Texas at Austin}
  \city{Austin}
  \state{Texas}
  \country{USA}
}

\author{Justin Chung}
\email{justin.chung@utexas.edu}
\authornotemark[1]
\affiliation{%
  \institution{The University of Texas at Austin}
  \city{Austin}
  \state{Texas}
  \country{USA}
}

\author{Stanley Chung}
\email{stanley.chung@utexas.edu}
\authornotemark[1]
\affiliation{%
  \institution{The University of Texas at Austin}
  \city{Austin}
  \state{Texas}
  \country{USA}
}
\author{Ranit Gupta}
\email{ranitgupta@utexas.edu}
\authornotemark[1]
\affiliation{%
  \institution{The University of Texas at Austin}
  \city{Austin}
  \state{Texas}
  \country{USA}
}

\author{Janani Ramamoorthy}
\email{janani.ram@utexas.edu}
\affiliation{%
  \institution{The University of Texas at Austin}
  \city{Austin}
  \state{Texas}
  \country{USA}
}

\author{Kartikeya Gullapalli}
\email{gkartikeyag@utexas.edu}
\affiliation{%
  \institution{The University of Texas at Austin}
  \city{Austin}
  \state{Texas}
  \country{USA}
}

\author{Diana Marculescu}
\email{dianam@utexas.edu}
\orcid{0000-0002-5734-4221}
\affiliation{%
  \institution{The University of Texas at Austin}
  \city{Austin}
  \state{Texas}
  \country{USA}
}

\author{Dimitrios Stamoulis}
\email{dstamoulis@utexas.edu}
\orcid{0000-0003-1682-9350}
\affiliation{%
  \institution{The University of Texas at Austin}
  \city{Austin}
  \state{Texas}
  \country{USA}
}

\renewcommand{\shortauthors}{Bhattaram et al.}

\begin{abstract}
We present GeoFlow, a method that automatically generates agentic workflows for geospatial tasks. Unlike prior work that focuses on reasoning decomposition and leaves API selection implicit, our method provides each agent with detailed tool-calling objectives to guide geospatial API invocation at runtime. GeoFlow increases agentic success by 6.8\% and reduces token usage by up to fourfold across major LLM families compared to state-of-the-art approaches.
\end{abstract}




%% information and the body of the document, and spans the page.
\begin{teaserfigure}
  \includegraphics[width=\textwidth]{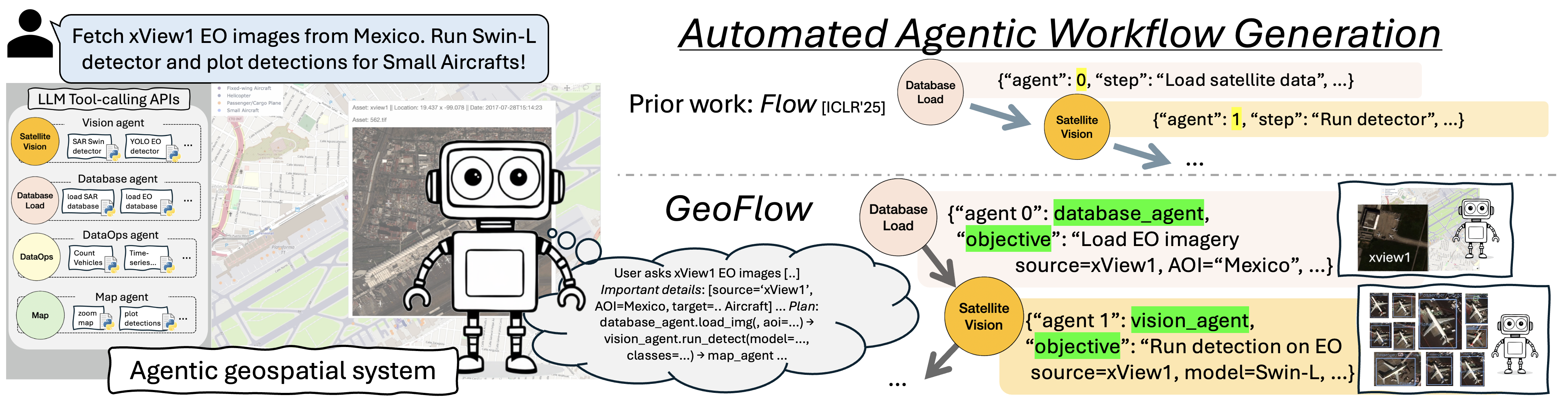}
  \caption{Automating agentic workflow generation for geospatial tasks with GeoFlow. The emerging paradigm of automatic workflow creation could hold tremendous potential for Earth Observation and remote sensing, but existing methods focus on reasoning decomposition while leaving API orchestration implicit (see Flow~\cite{niu2025flow}, top-right). In this study, we evaluate automatic workflows in geospatial tasks and introduce GeoFlow, which embeds explicit function-calling GIS API objectives (bottom-right) in agentic workflows represented as Activity-on-Vertex (AOV) graphs. GeoFlow improves agentic task success by 6.8\% over prior work while cutting token costs up to fourfold. Our full investigation is available at: \href{https://github.com/dstamoulis/geo-olms}{\textcolor{magenta}{https://github.com/dstamoulis/geo-olms}}.}
  \Description{Overview of proposed methodology.}
  \label{fig:teaser}
\end{teaserfigure}

\maketitle

\section{Introduction: Study Overview}

Recent advances in agentic AI have produced powerful multi‐agent frameworks. Empowered by large language models (LLMs), multiple LLM-based agents collaborate and execute tasks sequentially, each responsible for a specific function. Prior work has demonstrated their potential to augment geospatial analysis in Earth Observation~\cite{chen2024geoagent, singh2024geollm}, remote sensing~\cite{lee2025multi}, and sustainability studies~\cite{stamoulis2025geoolm}. However, existing multi-agent systems depend on manually designed workflows, \textit{i.e.}, predefined sequences of LLM invocations~\cite{wu2024stateflow}, to specify how agents should interact at runtime. This hinders their scalability and adaptability, especially on GIS platforms, where the types of geospatial tasks and the breadth of API tools involved in manual workflow construction can vary dramatically~\cite{stamoulis2025geoolm}.

An emerging paradigm in agentic AI aims to \textit{automatically} generate workflows by determining subtask allocations and roles in real time. Novel techniques such as Flow~\cite{niu2025flow}, AFlow~\cite{zhang2025aflow}, AutoFlow~\cite{li2024autoflow}, and MaaS~\cite{zhang2025maas} formulate agentic workflows as Activity-on-Vertex (AOV) graphs, \textit{i.e.}, directed acyclic graphs (DAG), where nodes represent subtasks with status and logs, and edges capture dependencies. To our knowledge, this novel paradigm has not been evaluated on geospatial tasks, thereby \textbf{motivating} our study. 

In this paper, we investigate state-of-the-art workflow automation methods to assess their effectiveness in executing Earth Observation tasks. We integrate Flow~\cite{niu2025flow} in an open-source geospatial agentic platform~\cite{stamoulis2025geoolm} and compare it to recent protocols, namely OpenAI's Swarm~\cite{openai2025swarm} and Microsoft AutoGen's ledger-based orchestration~\cite{wu2024autogen}. Our key \textbf{finding} is that Flow methods encounter challenges when subtasks require extensive API tool calls. As illustrated in Figure~1 (top right), we observe that Flow AOV subtasks might lack the specificity required to correctly identify the corresponding GIS tools. Prior work circumvents this issue by manually assigning agents at each workflow step~\cite{wu2024stateflow}, an assumption that limits scalability to geospatial systems with hundreds of API tools, mapping routines, and satellite product databases~\cite{singh2024geollm}.

Building on our observations, we introduce GeoFlow, a simple yet effective extension of the AOV formulation: during workflow generation at runtime, GeoFlow \textit{explicitly} assigns function-calling objectives and the corresponding APIs to subagents (Figure~1, bottom right). Our results show that our method outperforms existing approaches, achieving 6.8\% higher task completion rates on the GeoLLM-Engine benchmark~\cite{singh2024geollm}. These gains hold across multiple model families, including OpenAI GPT, Qwen, Mistral, and Llama. To support future research on automated workflows for GIS tasks, our full implementation is available on our project repo.

\section{Methodology}

\textbf{Background.} Agentic automation methods represent workflows as AOV graphs in which vertices denote subtasks and edges denote precedence relations, following a two-stage pipeline: (1) workflow generation and (2) execution. First, given a user task $T$, a meta-agent LLM generates the dictionary-based graph $G$, where each vertex (subtask) is assigned to a subagent. Then, by traversing $G$ at runtime, each subagent relies on the shared global chat (LLM messages) history to infer which API tools to invoke and when to return control back to the system. This works for general-knowledge tasks (\textit{e.g.}, trivia quizzes or math problems) where each subtask trivially maps to a distinct API (\textit{e.g.}, calculator \textit{vs}. wiki). However, GIS platforms introduce far more complexity, as illustrated next.

Consider an AOV with subtasks ``\textit{load satellite data}'' and ``\textit{run detector}'' (Fig.~1, top-right): Flow leaves it to the \texttt{vision\_agent} to deduce the exact satellite vision model based on global history and the preceding messages by the \texttt{database\_agent}. Our investigation shows (Section~\ref{sec:results}) that not specifying such reasoning aspects for subsequent steps causes subagents to invoke the wrong functions, especially when smaller LLMs are exposed to multiple API tools to choose from (\textit{e.g.}, different satellite vision models trained on EO versus SAR imagery for various ground-object categories).

\vspace{+5pt}\noindent
\textbf{GeoFlow.} In this study, we investigate a simple extension to the AOV formulation: at workflow generation, GeoFlow is prompted to populate each vertex with a precise ``\textit{agentic scope and objective}'' containing detailed instructions. For example, instead of ``\textit{run detector},'' the objective would read: ``\textit{The database agent should be providing you with loaded EO imagery for dates 2024-XX to YY over AOI Z; run the Swin-L EO detector and return class A.}'' By fully contextualizing inter-agent dependencies and required operations, our workflow gives subagents the context they need to invoke the correct geospatial APIs. Formally, we denote the AOV graph $G = \left(V, E, A, O\right)$, where $V$ is the set of all subtasks (vertices); $E$ is the set of directed edges indicating subtasks dependencies, $A$ is the set of agents, and $O$ maps each vertex to a concrete agentic objective.

\begin{figure*}[t]
  \centering
  \includegraphics[width=\linewidth]{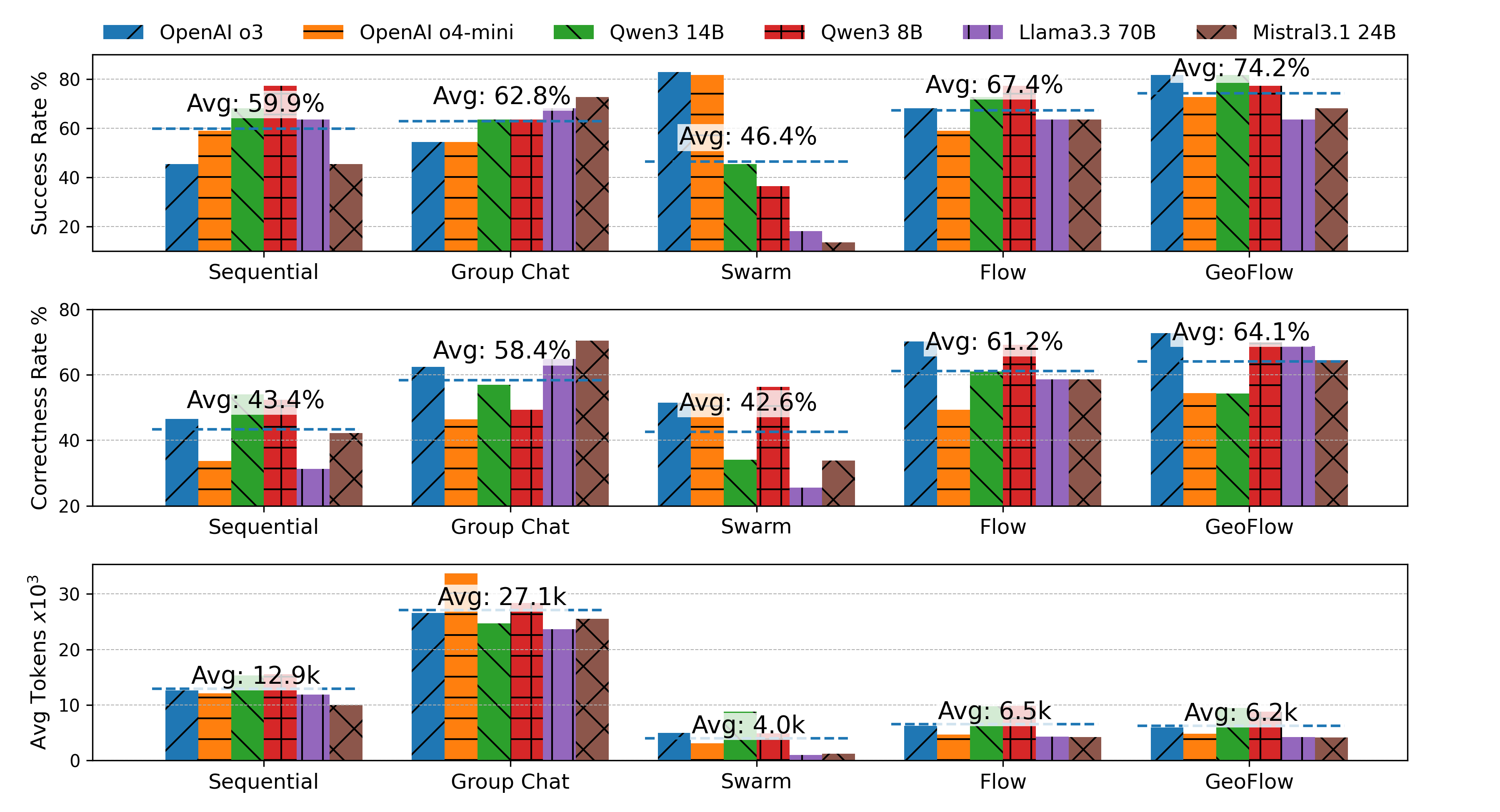}
  \vspace{-10pt}
  \caption{GeoFlow outperforms state-of-the-art multi-agent designs on 20 GeoLLM-Engine tasks~\cite{singh2024geollm} in success rate (top), correctness rate (middle), and token usage (bottom). Our approach improves average task success by 6.8\% and 2.9\%, respectively, compared with Flow~\cite{niu2025flow}, while reducing token cost by over $4\times$ relative to AutoGen Group-Chat~\cite{wu2024autogen}.}
  \label{fig:barplot_main}
  \Description{Agentic performance.}
\end{figure*}

\begin{tcolorbox}[
  enhanced,
  colback=gray!30!white,
  colframe=gray!70!black,
  coltitle=white,
  title=\bfseries Sample prompt: Geospatial workflow generation,
  fonttitle=\large\bfseries,
  sharp corners=south,
  attach boxed title to top center={yshift=-2mm},
  boxed title style={
    colframe=gray!70!black,
    colback=gray!70!black,
  }
]
\small
\ttfamily
You are a workflow planner for geospatial tasks [..] choosing only from the following GIS agents [..]:
\vspace{1ex}

- database\_agent: APIs fetching satellite images [..]

- vision\_agent: satellite vision APIs [..] ...

\vspace{1ex}
Valid output format example:
\{
    tasks: \{
        "task0": \{
          "id": "task0",
          "objective": "Load satellite imagery from ..",
          "next": [
            "task1"
          ],
          "prev": [],
          "status": "pending",
          "agent": "database\_agent"
        \},
        "task1": \{
          "id": "task1",
          "objective": "Run vision model ..",
          "next": [
            "task2"
          ],
          "prev": [
            "task0"
          ],
          "status": "pending",
          "agent": "vision\_agent"
        \}
    \}
\}
\end{tcolorbox}

% \vspace{+5pt}
\noindent
\textbf{Workflow generation}. Given a geospatial task $T$, the meta-agent LLM is prompted to generate the workflow graph $G$ in the dictionary-based format of~\cite{niu2025flow} (sample above). The LLM is supplied with descriptions of the agents in our geospatial platform and their available APIs. Unlike Flow, where the agent set $A$ serves as ordering indexing (\texttt{agent\_0}, \texttt{agent\_1}, $\dots$) and API matching is left implicit, we require the LLM to return each agent with its designated API name. The LLM then populates the objectives $O$ with all GIS study parameters -- area of interest, target time range, data source, and specific map operations. A JSON dictionary output is generated, which is fed into the geospatial system as an execution trace.

\vspace{+5pt}
\noindent
\textbf{Workflow execution}. Once the meta-agent produces the AOV graph, the execution order is calculated using a topological sort on its indexed vertices, as seen in Flow~\cite{niu2025flow}. For each subtask $v$ in the sorted sequence, we invoke the corresponding tool-augmented LLM subagent $a$ via standard function-calling. Each subagent uses its API tools to complete the objective and returns results (\textit{e.g.}, loaded datasets or vision detection outputs) that feed into downstream subtasks. The meta-agent tracks progress through the global chat history and reports overall task completion to the user. If an error occurs, control returns to the meta-agent, which uses the chat history and error message to refine $G$ via update prompts~\cite{niu2025flow}.
\section{Experimental Setup}

\noindent
\textbf{Geospatial platform.} We conduct our experiments within the Geo-OLM~\cite{stamoulis2025geoolm} platform, an open-source implementation of GeoLLM-Engine that integrates non-proprietary (Ollama) LLMs. Following the official Flow codebase, we implement both Flow and GeoFlow. Moreover, we integrate two AutoGen orchestration designs: Group-Chat and Sequential~\cite{wu2024autogen}. Last, we consider the OpenAI Swarm, which emulates agent handoffs via LLM function-calling. We test all methods using the latest variants of major LLM families, including OpenAI o4-mini and o3, Mistral 3.1, Llama 3.3, and Qwen 3.

\vspace{+5pt}\noindent
\textbf{Benchmark and metrics.} We evaluate GeoFlow on the GeoLLM-Engine benchmark~\cite{singh2024geollm}, which comprises realistic Earth Observation tasks for object detection and land cover classification over millions of satellite images. Replicating the Flow setup~\cite{niu2025flow}, we select 20 benchmark queries, construct ``ground truth'' AOV, and programmatically generate function-calling execution traces. To ensure representative comparisons, an additional ``oracle'' example is tested using few-shot prompting across all methods (Swarm, AutoGen Group-Chat, Sequential, Flow, and GeoFlow). Agentic performance is assessed using \textit{success} and \textit{correctness} rates. Success rate is the proportion of fully completed tasks, regardless of intermediate errors, while correctness rate is the fraction of correct API tool calls~\cite{paramanayakam2025less}.

\section{Results}
\label{sec:results}

\noindent
\textbf{Overall results.} GeoFlow achieves the highest average performance across both success rate (74.2\%) and correctness rate (64.1\%) among all evaluated methods (Fig.~\ref{fig:barplot_main}). Compared to the closest baseline, Flow, we improve agentic success by 6.8\% (74.2\% \textit{vs}.\ 67.4\%) and correctness by 3.3\% (64.1\% \textit{vs}.\ 61.2\%), validating our hypothesis that incorporating explicit objectives improves agentic function-calling. In contrast, AutoGen baselines (Sequential and Group Chat) exhibit considerably lower correctness. In our experiments, we observed frequent cases where agents in these baselines attempted actions beyond the task scope due to ambiguity in return conditions. Notably, Swarm performs competitively with OpenAI models, suggesting strong alignment with their internal handoff protocols. However, its performance degrades sharply with open-source models.

\vspace{+5pt}\noindent
\textbf{Model families.} GeoFlow maintains robust agentic performance with non-proprietary LLM models. We note that Qwen models, known for their strong function-calling capabilities, achieve 77.3\% success and 70.0\% correctness, consistent with prior geospatial API evaluations~\cite{stamoulis2025geoolm}. Interestingly, we observe degraded performance when using OpenAI models in the Sequential method. In our experimentation, we observed that these models would often ``over-reason'' and revise plans, eventually trying to erroneously invoke tools beyond their scope (tool set).

\vspace{+5pt}\noindent
\textbf{Performance \textit{vs}. cost trade-off.} 
GeoFlow shows the best trade-off between performance and cost among all evaluated methods. Compared to Flow, the second-best performing baseline, GeoFlow achieves higher agentic rates at nearly the same token usage (6.2k \textit{vs}.\ 6.5k). When comparing to AutoGen Group Chat, the strongest non-Flow baseline, GeoFlow achieves significant improvement in average success rate (74.2\% \textit{vs}. 62.8\%), while requiring 4$\times$ fewer tokens on average (6.2k \textit{vs}. 27.1k). This is expected, as orchestration-based approaches like Group Chat rely on consensus coordination between the central meta-agent and subagents, incurring substantial communication overhead.

\begin{table}[t!]
  \caption{Workflow correctness based on LLM-Scores~\cite{majumdar2024openeqa}.}
  % \vspace{-5pt}
  \label{tab:flow_scores}
  \centering
  \begin{tabular}{l c}
    \toprule
    \textbf{Model} & \textbf{AOV $G$. Average Flow Score}  \\
    \midrule
    OpenAI o3 & 96.61\% \\
    OpenAI o4-mini & 96.01\% \\
    Qwen3 14B & 97.73\% \\
    Qwen3 8B & 95.66\% \\
    Llama3.3 70B & 94.86\% \\
    Mistral3.1 24B & 96.31\% \\
    \bottomrule
  \end{tabular}
\end{table}

\vspace{+5pt}\noindent
\textbf{Workflow ``correctness'' ablation.} To assess the quality of generated AOVs $G$, we adapt the existing correctness metric -- as originally defined over sequence of function-calls -- by performing a depth-first search (DFS) on $G$ and grouping steps by API agent (\textit{e.g.}, the sequence of all subtasks assigned to \texttt{database\_agent} ). We apply the LLM-Score technique~\cite{majumdar2024openeqa} that evaluates LLM generated objectives: we prompt GPT-4o to assign a score from 1 (poor) to 5 (perfect) for the objectives against the manually crafted ground truths. We count an error whenever there is a structural $G$ mismatch (missing edges or vertices) or when the objective LLM-Score is below 4. We report the average ``Flow Score'' over the 20 geospatial tasks considered (Table~\ref{tab:flow_scores}): all models achieve close to 95\% AOV ``correctness''. We note, however, that GeoLLM-Engine tasks exhibit mostly linear logic and dependencies. As future work, we motivate evaluating workflow generation on more complex, interactive multi-round tasks in which a GIS analyst actively updates the system environment (\textit{e.g.}, maps and databases).

\begin{figure}[h!]
  \centering
  \includegraphics[width=\linewidth]{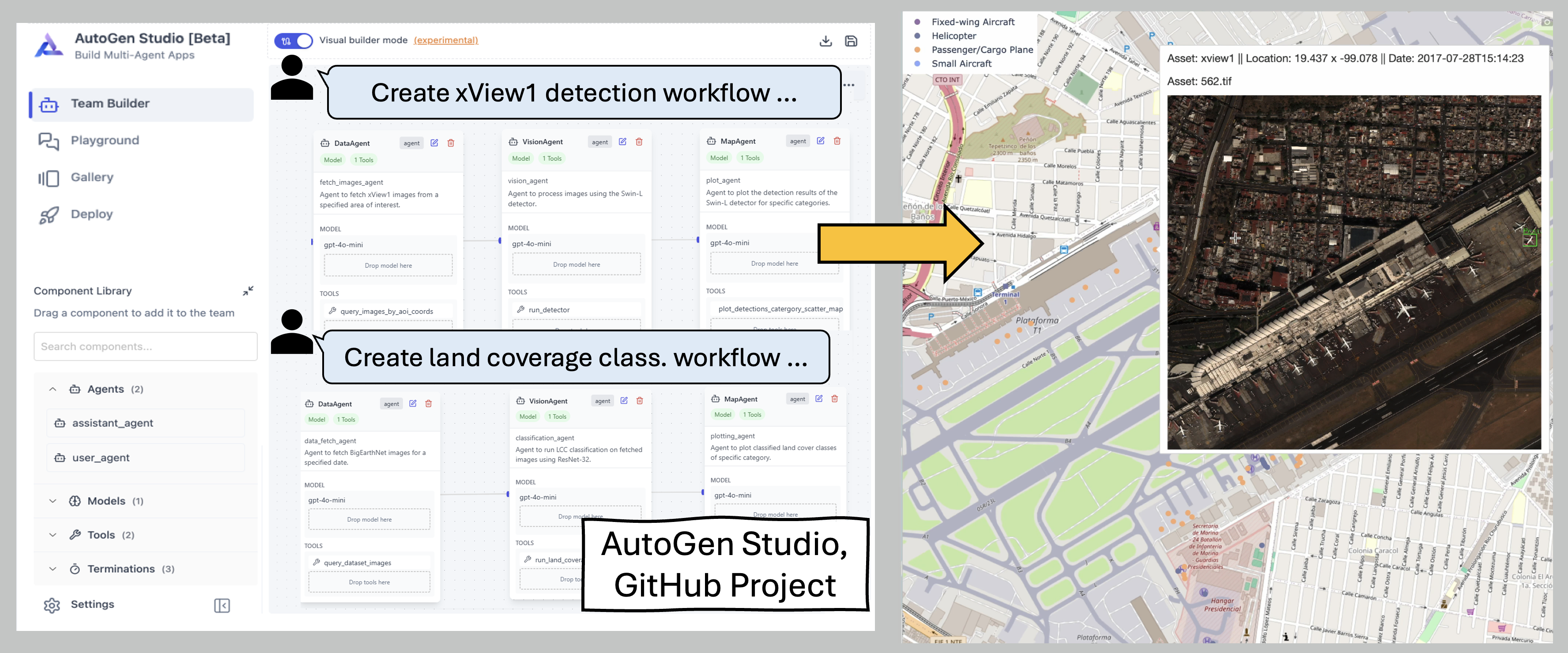}
  \vspace{-5pt}
  \caption{Workflow visualizations rendered using the open-source AutoGen Studio UI tool~\cite{dibia2024autogenstudio} (left). We motivate the potential of human-in-the-loop geospatial agentic systems: users can update AOV-generated workflows in a no-code UI~\cite{dibia2024autogenstudio} and execute them directly in GIS platforms~\cite{stamoulis2025geoolm} (right).}
  \label{fig:ui_idea}
  \Description{Using automatic flow generation in GIS UIs.}
\end{figure}

\section{Discussion and Future Work}

We highlight an overlooked aspect in current workflow automation research: its potential as an agentic \textbf{no-code} design tool that would allow GIS practitioners to build and refine multi-agent pipelines without agentic AI expertise. In our investigation, we saw that meta-LLM-generate AOV graphs generated can serve not only as execution traces but also as editable designs in a human-in-the-loop interface. Unlike existing no-code workflow systems (\textit{e.g.}, Azure Prompt flow) that still demand manual flow construction, this paradigm would allow users to visualize, adjust agent assignments and dependencies, and then execute the finalized workflow directly in a GIS system. In Figure~\ref{fig:ui_idea}, we illustrate this concept using AutoGen Studio UI~\cite{dibia2024autogenstudio} to render example geospatial workflows before passing them to our geospatial platform. We believe that integrating automated graph generation with interactive UI controls represents a promising direction for GIS-focused HCI research and we will explore it in future work.

\section{Conclusion} 
In this paper, we presented GeoFlow, an extension of Flow's AOV workflow generation that embeds explicit function-calling objectives and API specifications for geospatial subagents. We integrated GeoFlow into Geo-OLM and benchmarked it against Flow, OpenAI Swarm, and AutoGen's multi-agentic designs across major LLM families on 20 GeoLLM-Engine tasks. Our evaluation showed that GeoFlow raised task success by 6.8\% over Flow, while reducing token usage by up to fourfold.

\bibliographystyle{ACM-Reference-Format}
\bibliography{geoflow}

\end{document}